\documentclass[a4paper,twoside]{article}
\pdfoutput=1
\usepackage{epsfig}
\usepackage{subcaption}
\usepackage{calc}
\usepackage{amssymb}
\usepackage{amstext}
\usepackage{amsmath}
\usepackage{amsthm}
\usepackage{multicol}
\usepackage{pslatex}
\usepackage{apalike}

\usepackage{soul}
\usepackage{graphicx}
\usepackage{url}
\usepackage{latexsym}
\usepackage{pdfpages}

\usepackage{booktabs} 
\usepackage{tabularx}
\usepackage{enumitem}
\usepackage[utf8]{inputenc} 
\usepackage{multirow} 
\usepackage{balance}
\usepackage{graphics}
\usepackage[bottom]{footmisc}

\usepackage{SCITEPRESS}     % Please add other packages that you may need BEFORE the SCITEPRESS.sty package.

\begin{document}

\title{On Informative Tweet Identification For Tracking Mass Events}

\author{\authorname{Renato Stoffalette Jo\~{a}o\orcidAuthor{0000-0003-4929-4524}}
\affiliation{L3S Research Center\\Leibniz University of Hannover\\ Appelstra{\ss}e 9A - Hannover\\ 30167, Germany}
\email{joao@L3S.de}
}


\abstract{Twitter has been heavily used as an important channel for communicating and discussing about events in real-time. In such major events, many uninformative tweets are also published rapidly by many users, making it hard to follow the events. In this paper, we address this problem by investigating machine learning methods for automatically identifying informative tweets among those that are relevant to a target event. We examine both traditional approaches with a rich set of handcrafted features and state of the art approaches with automatically learned features. We further propose a hybrid model that leverages both the handcrafted features and the automatically learned ones. Our experiments on several large datasets of real-world events show that the latter approaches significantly outperform the former and our proposed model performs the best, suggesting highly effective mechanisms for tracking mass events.}

\onecolumn \maketitle \normalsize \setcounter{footnote}{0} \vfill

\section{\uppercase{Introduction}}
\label{introduction}
\vspace{-2mm}
%\noindent

Lately Twitter has become an important channel for communication and information broadcasting. A large number of its users have been using the platform for seeking and sharing the information about events. Particularly, during undesired mass events like natural disasters or terrorist attacks, Twitter users post tweets, share updates, inform other users about current situations, etc. However, in addition to these information, a lot of tweets are merely for discussing and expressing opinions and emotions towards the events, which makes it challenging for professionals involved in crisis management  to actually collect relevant information for better understanding the situations and respond more rapidly \cite{vieweg2010microblogging}.

Considering the large volume of tweets published by Twitter users, manual sifting to find useful information is inherently impractical \cite{meier2013crisis}. Thus automatic mechanisms for identification of the informative tweets are required to assist not only the average citizen to become aware of the situation but also the professionals to take measures immediately and potentially save lives.
%\begin{table*}
%    \small
%    \centering
%    \caption{Examples of tweets from \textsc{CrisisMMD}\cite{crisismmd2018icwsm} dataset.}
%    \begin{tabular}{||m{35em} c||}  \hline
%         \textbf{Tweet }&  \textbf{Label}\\ \hline  \hline 
        %\textit{\#SriLanka floods: 206 people dead, 92 still missing https://t.co/goLNqtiZUX \#top \#news https://t.co/jJ9YCNSL4S} & \textit{Informative} \\ \hline
        %\textit{Thousands Homeless as Mexico Quake's Death Toll Tops 300 https://t.co/4iSf2hMv4m https://t.co/t28wYIcQoY}  & \textit{Informative}\\ \hline
%        \textit{CR 218 bridge is closed, after Black Creek flooded during Hurricane Irma. Live at 5. @ActionNewsJax https://t.co/MDbNr7HnTh}  & \textit{Informative}\\ \hline
        %\textit{Glad to be alive. \#lincoln \#ford \#Garmin \#mkx \#geico \#HurricaneHarvey https://t.co/rgNfWHcnxo} & \textit{Not informative} \\ \hline
        %\textit{@insideFPL it's been almost been 10 days.Please keep your promise. \#frustrating \#irma \#fpl https://t.co/ISapDMh5Vl} & \textit{Not informative} \\ \hline
%        \textit{i love huge murals!!!!!!!!! (by izak walter mora marambio) https://t.co/X4tPIG995y}  &\textit{ Not informative} \\ \hline
%    \end{tabular}
%    \label{tab:tweetsExample}
%\end{table*}

In this work, we investigate the viability of machine learning approaches for developing such an automatic mechanism. We study both traditional ones that use handcrafted features, as well as the state of the art representation learning approach, the BERT-based models \cite{devlin2018bert},  to classify tweets according to their informativeness. %Examples of \textit{Informative} and \textit{Not Informative} tweets from the CrisisiMMD dataset ~\cite{crisismmd2018icwsm} are demonstrated in Table~\ref{tab:tweetsExample}.
We implement a rich set of features for the former, examine different usage of the latter as well the combinations of both. Furthermore, we propose a hybrid model that leverages both the BERT-based models and the handcrafted features. We evaluate all these models on large datasets collected during several natural and man-caused disasters. In summary, we make the following contributions.
\begin{itemize}

    \item We investigate a rich set of features that include Bag-of-Words, text-based, and user-based features for traditional models, and examine the performance of BERT-based models for the informative tweet classification problem.
    
    \item We further propose a hybrid model that combines a BERT-based model with handcrafted features for the problem.
    
    \item We conduct comprehensive experiments for evaluating the performance of these diverse models.
    
    \item Empirically, we demonstrate that deep BERT-based models outperform the traditional ones for the task without requiring complicated feature engineering, while our proposed model performs the best. 
\end{itemize}

The remaining of this paper is organized as follows. We firstly review the related works in Section \ref{motiv_rw}, then we describe the methods and the features in Section \ref{methodology}. Section \ref{experiments} describes our experiments, datasets and give details about our implementation methods. In Section \ref{results} we report the results obtained from our experiments. Finally, we draw some conclusions and point out some future directions in Section \ref{Conclusion}.

\section{\uppercase{Related Work}}
\label{motiv_rw}
\vspace{-2mm}

Social media platforms such as Twitter and Facebook have become valuable communication channels over the years. Twitter enables people to share all kinds of information by posting short text messages, called tweets. Although social media services are full of conversational messages, it is also an environment where users post newsworthy information related to some natural or human-induced disaster. Identifying such information can help not only the ordinary citizen but it can also assist professionals and organizations in coordinating their response for potentially saving lives and diminishing catastrophic losses \cite{imran2015processing}.

A number of automated systems have been proposed to extract and classify crisis related information from social media channels, for example
CrisisTracker~\cite{rogstadius2013crisistracker}, Twitcident~\cite{abel2012semantics}, AIDR~\cite{imran2014aidr}, among others. For a more complete list of systems, please refer to the survey by Imran et al.~\cite{imran2015processing}. 

Machine learning and natural language processing play an important role when it comes to classifying crisis related tweets automatically, and the approach applied to extract textual features can determine the performance of an automated classifier.
Castillo et al.  \cite{castillo2011information} proposed automatic techniques to assess the credibility of tweets related to specific topics or events, using features extracted from user's posting behavior and tweet's text.
Verma, et al.~\cite{verma2011natural} used Naive Bayes and MaxEnt classifiers to find situational awareness tweets from several crises and Cameron et al.~\cite{cameron2012emergency} described a platform for emergency situation awareness where they classified interesting tweets using an SVM classifier.

With the recent advances in natural language processing and the emergence of techniques such as word2vec \cite{mikolov2013efficient,mikolov2013distributed} and GloVe \cite{pennington2014glove}, deep neural networks have successfully been applied in similar tasks. Caragea et al. \cite{caragea2016identifying}  for instance, demonstrated that convolutional neural networks outperformed traditional classifiers in tweet classification. Nguyen et al.~\cite{nguyen2017robust} also used a convolutional neural network based model to classify crisis-relevant tweets. These results suggest a promising approach for this informative tweet classification task.

\section{\uppercase{Methodology}}
\label{methodology}
\vspace{-2mm}

Identifying informative tweets is a critical task, particularly during catastrophic events. There is however no simple rules that can be applied for the task. We therefore approach the problem of informative tweets identification as a supervised learning problem. 
In the following subsections, we shall discuss several models for the task. We start with some conventional classification models that make use of features engineered from the tweets as well as the users who posted the tweets. Next, we present the deep learning approaches for the task, and describe our proposed model.

\subsection{Traditional models}
\vspace{-2mm}

Several machine learning approaches have been proposed for the task of automatically detecting crisis-related tweets, for example, Naive Bayes \cite{li2018disaster},
Support Vector Machines \cite{caragea2016identifying}, and Random Forests \cite{kaufhold2020rapid}.
Thus, as the baselines, we have trained these traditional classifiers to automatically classify a tweet into either \textit{Informative} or \textit{Not Informative}. Specifically, we have implemented the following models.
\begin{itemize}
    \item \textsc{\textbf{Logistic Regression (LR)}} - a classifier that models the probability of a label based on a set of independent features, \item \textsc{\textbf{Decision Tree (DT)}} - a classifier that successively divides the features space to maximise a given metric (e.g., information gain), 
    \item \textsc{\textbf{Random Forest (RF)}} - a classifier that utilises an ensemble of uncorrelated decision trees, 
    \item \textsc{\textbf{Naive Bayes (NB)}} - a Gaussian Naive Bayes classifier,
    \item \textsc{\textbf{Multilayer Perceptron (MP)}} - a network of linear classifiers, (\textit{perceptrons}) that uses the backpropagation technique to classify the instances, and
%\item \textbf{Adaboost (AB):} Adaboost or Adaptive Boosting is an iterative ensemble boosting classifier that combines multiple classifiers to increase the accuracy of poorly performing classifiers.
    \item 
    \textsc{\textbf{Support Vector Machine (SVM)}} - a discriminative classifier formally defined by a separating hyperplane. %In other words, given labeled training data (supervised learning), the algorithm outputs an optimal hyperplane which categorizes new examples. In two dimentional space this hyperplane is a line dividing a plane in two parts where in each class lay in either side.
\end{itemize}

All the classifiers deployed in this work were implemented in Python using the machine learning library Scikit-Learn \cite{pedregosa2011scikit}.
The source code of our models implementations is freely available at \url{https://github.com/renatosjoao/infotweets.git}.

\subsubsection{Features}
\label{lab:features}
\vspace{-2mm}

Inspired by previous works, we investigated a set of features based on the tweets' contents as well as on the users who posted the tweets \cite{acerbo2017filtering,graf2018cross,imran2013extracting,verma2011natural}. These features are described as follows.

\begin{itemize}
    \item Text-based features: the ones that are calculated from the content of a tweet, including
        \begin{itemize}
            \item \textbf{$n_{chars}$:} This feature refers to the number of characters a tweet contains.

            \item \textbf{$n_{words}$:} The number of words a tweet contains. After removing symbols and patterns we count the number of words that is present in the tweet.

            \item \textbf{$n_{hashtags}$:} The number of occurrences of \#hashtags in a tweet. It  can indicate the user wants to highlight some specific subject of interest. 

            \item \textbf{$n_{url}$:} The number of URLs contained in a tweet. 

            \item \textbf{$n_{at}$:} The number of \textit{@} tags in the tweet can be an indicator that the user is tagging people to draw their attention.

            \item \textbf{$b_{hashtag}$:} Binary valued feature referring to the presence of \#hashtags. True if at least one \#hashtag is present in the tweet, false otherwise.

            \item \textbf{$b_{at}$:} Binary valued feature referring to the presence of @ tags. True if the tweet contains @ tags, false otherwise.

            \item \textbf{$b_{rt}$:} Binary valued feature referring to a retweeted message. True if the tweet contains retweet patterns, such as rt@, false otherwise.

            \item \textbf{$b_{slang}$:} Binary valued feature referring to slangs in the tweet. True if the tweet contains any slang, false otherwise. Internet abbreviations are examples of text informality, which are representative of conversations. We built a dictionary of slangs from an online slang dictionary\footnote{https://www.lifewire.com/urban-internet-slang-dictionary-3486341}.

            \item \textbf{$b_{url}$:} Binary valued feature about the presence of URLs. True if at least one URL is present in the tweet, false otherwise.

            \item \textbf{$t_{lex}$:} Tweet lexical diversity refers to the number of unique words divided by the total number of words in the tweet.

            \item \textbf{$b_{interj}$:} Binary valued feature referring to an interjection. True if the tweet contains interjections, false otherwise. We built a dictionary of interjections from an online list of interjections\footnote{https://www.vidarholen.net/contents/interjections/}.

            \item \textbf{$bow$:} Bag-of-Words features. Real-valued vectors are calculated with TF$\times$IDF of the words and Twitter posts from each corpus for a finite number of words from the vocabulary.\\
        \end{itemize}
        \item User-based features: the ones that are calculated from the user who posted the tweet, including
        \begin{itemize}
            \item \textbf{$b_{usr}$:} Binary valued feature representing whether the user account is verified. True if the  user has a verified account, false otherwise.

        \item \textbf{$n_{followers}$:}  This feature represents the number of followers the user who posted the tweet has. Since this number may vary considerably we calculated it as $\textit{log}_{10}(n_{followers}+1)$.

        \item \textbf{$n_{followees}$:} Number of accounts the user who posted the tweet follows, calculated as $\textit{log}_{10}(n_{followees}+1)$.

        \item \textbf{$n_{tweets}$:} This feature represents the total number of tweets posted by the user. There can be the case where the user has not posted many tweets as well as there can be cases of influential users who post messages more frequently, thus we calculate this feature as $\textit{log}_{10}(n_{tweets}+1)$.
        \end{itemize}
    \end{itemize}

\subsection{Deep learning approaches}
\vspace{-2mm}

We now discuss deep learning based approaches that are widely used in recent works \cite{nguyen2017robust,neppalli2018deep}.

\subsubsection{Word embedding methods}
\vspace{-2mm}

The traditional models such as the \textit{Bag-of-Words} do not capture well the meaning of the words and consider each word as a separate feature. Word embeddings have been proposed and widely used neural models that map words into real number vectors such that similar words are closer to each other in a higher dimensional space. The word embeddings captures the semantical and syntactical information of words taking into consideration the surrounding context. 

In this work, we examine the following typical word embedding methods:
\begin{itemize}
    \item \textbf{Word2vec} \cite{mikolov2013distributed} is one famous method of neural words embeddings initially proposed in two variants: (i)  a Bag-of-Words model that predicts the current word based on the context words, and (ii) a skip-gram model that predicts surrounding words given the current word.

    \item \textbf{GloVe} is an extension to the Word2vec method for efficiently learning word vectors, proposed by \cite{pennington2014glove} which uses global corpus statistics for words representations and learns the embeddings by dimensionality reduction of the co-occurrence count matrix. 
    \item \textbf{Fasttext}  \cite{bojanowski2016enriching} is an extension to the skip-gram model from the original Word2vec model which takes into account subword information, i.e. it learns representations for character n-grams, and represents words as the sum of the n-gram vectors. The idea is to capture morphological characteristics of words.
\end{itemize}

We make use of the pre-trained word vectors of the above models\footnote{Word2vec: https://code.google.com/archive/p/word2vec/},\footnote{GloVe: https://nlp.stanford.edu/projects/glove/},\footnote{Fasttext: https://fasttext.cc/}. The feature vector of each tweet is then determined by taking the average of all embedding vectors of its words.

\subsubsection{Text embedding methods}
\vspace{-2mm}

Generalized from word embeddings, text embedding methods compute a vector for each group of words taken collectively as a single unit, e.g., a sentence, a paragraph, or the whole document. In this work, we examine a typical method for text embedding, namely Doc2vec, and state-of-the-art ones, namely BERT-based models.

{\bf Doc2vec} generates efficient and high quality distributed vectors of a complete document \cite{mikolov2013distributed}. The main objective of Doc2Vec is to convert the sentence (or paragraph) into a vector. It is a generalization on the Word2vec model.

{\bf BERT} is a model developed on a multi-layer bidirectional Transformer encoder \cite{vaswani2017attention,devlin2018bert}. It makes use of an attention mechanism that learns contextual relations between words in texts.
In its generic format, the Transformer includes two separate mechanisms, an encoder that reads input text and a decoder that produces the task prediction. 
The encoder is composed of a stack of multiple layers, and each layer has two sub-layers. The first is a multi-head self-attention mechanism, and the second is a simple, position-wise fully connected feed-forward network.
The decoder is also composed of a stack of multiple identical layers with the addition of a third sub-layer, which performs multi-head
attention over the output of the encoder stack.
One key component of the Transformer encoder is the multi head self-attention layer, i.e. a function that can be formulated as querying a dictionary with key-value pairs.

The most straightforward usage of BERT is to employ it as a blackbox for feature engineering. This is the combination of the default BERT model and conventional classifiers. The final hidden state of the first word ([CLS]) from BERT is the encoded sentence representation and it is input to conventional classifiers for the predictions task.

The original BERT model is pre-trained in a general domain corpus. Thus, for a text classification task in a specific domain, the data distribution may be different. In this way in order to obtain improved results, we need to further train BERT on a domain specific data. There are a couple of ways to further train BERT on a domain specific corpus. The first one is to train the entire pre-trained model on the new corpus and feed the output into a softmax function. In this way, the error is back propagated throughout the entire model's architecture and the weights are updated for this domain specific corpus. Another method is to train some of BERT's layers while freezing others, or we can freeze all the layers and attach extra neural network layers and train this new model where only the weights of the attached layers will be updated. These are so called fine tuning procedures, and in this work we will be fine tuning BERT, by encoding Twitter sentences with the BERT encoder and running more training iterations and backpropagating the error throughout the entire model. 

\subsection{Our proposed model}
\label{ourmodel}
\vspace{-2mm}

We now describe a hybrid model, called \textsc{BERT}$_{Hyb}$, that combines both the handcrafted features with the ones learned by BERT.
%Figure \ref{fig:modelarchitecture} shows an overview of our model architecture: 
\textsc{BERT}$_{Hyb}$ model feeds a vector of handcrafted features from the tweet through a linear layer, and also feeds the vector produced by BERT for the first token (CLS) of the tweet through another linear layer. The outputs of these two layers are concatenated and fed through a third linear layer, whose output is subsequently fed through a softmax layer to produce the prediction whether a tweet is \textit{Informative} or \textit{Not Informative}.
\section{\uppercase{Experiments}}
\label{experiments}
\vspace{-2mm}

We now present our experiments to empirically evaluate the methods presented above. In the following subsections, we shall describe the datasets, define the evaluation metrics, the experiment settings, and report the results.

\subsection{Datasets}
We use the following datasets to evaluate the models.
\begin{itemize}
    \item \textbf{\textsc{CrisisLexT26}} \cite{olteanu2015expect} - This is a dataset of  tweets collected during twenty six large crisis events in 2012 and 2013, with about 1,000 tweets labeled per crisis for informativeness, information type, and source. 
    %In this dataset, for each tweet there is an informative label, which can assume four different values: \textit{Related and Informative}, \textit{Related}, \textit{Not Related} and \textit{Not Applicable}. Since our focus is the evaluation of tweets informativeness, we exclude all tweets labeled as \textit{Not Related} or \textit{Not Applicable}.
    \item \textbf{\textsc{CrisisLexT6}} \cite{olteanu2014crisislex} - This dataset includes English tweets posted during six large events in 2012 and 2013, with about 60.000 tweets labeled by relatedness as \textit{On-topic} or \textit{Off-topic} with each event. We assume the tweets labeled as \textit{On-topic} being the \textit{Informative} tweets and \textit{Off-topic} being \textit{Not Informative} respectively.
 
    \item \textbf{\textsc{CrisisMMD}} \cite{crisismmd2018icwsm} - CrisisMMD is a dataset that contains tweets with both text and image contents.
    There are 16,000 tweets that were collected from seven events that took place in 2017 in five countries. 
   
    \item \textbf{\textsc{Covid}} \cite{covid19tweet} -  This dataset consists of 10K English Tweets collected during the Covid pandemic. It is split into training set with 3303 \textit{Informative}  tweets and 3697 \textit{Uninformative} tweets, and a validation set with 472 and 528 \textit{Informative} and \textit{Uninformative} tweets respectively.
    
\end{itemize}

In their original form, the above datasets provide only tweets' content together with their ids and labels. To calculate the user based features we crawl from Twitter the full information of all the tweets. However, some tweets are no longer available. We thus create a version of each dataset that consists of the subset of tweets that we can crawl full information from Twitter. These versions are \textsc{Covid} and \textsc{Covid}$_{SUBSET}$, \textsc{CrisisLexT6} and \textsc{CrisisLexT6}$_{SUBSET}$,  \textsc{CrisisLexT26} and \textsc{CrisisLexT26}$_{SUBSET}$, \textsc{CrisisMMD} and \textsc{CrisisMMD}$_{SUBSET}$ respectively. The basic statistics of all the datasets and their subsets are shown in Tables~\ref{tbl:datasetsstats} and~\ref{tbl:datasetsstatsSubset} respectively.

\begin{table*}[t]
\centering
\caption{Complete datasets classes distributions.}
%\vspace{-2.5mm}
\label{tbl:datasetsstats}
\small
\begin{tabular}{|l|r|r|r|}	
     \toprule
     \multicolumn{1}{|c|}{\textit{Dataset}}    & \multicolumn{1}{|c|}{\textit{\#Informative}}    & \multicolumn{1}{|c|}{\textit{\#Not Informative}}     & \multicolumn{1}{|c|}{\textit{Total}}   \\ 
     \midrule
      \textsc{Covid}     &    3,772 & 4,221 & 7,993 \\
	  %- - - Train & 3300 & 3694  & 6994\footnote{removed duplicates}  \\
	  %- - - Val   &  472 &  527  & 999\footnote{removed duplicates}  \\
      %- - - Test  &  944 & 1056  & 2000\\ labelled test has not being released yet
      %- - - TOTAL  & 3772 & 4221 & 7993\footnote{removed duplicates} \footnote{Missing the labelled test set}\\
     \textsc{CrisisLexT6}   &    32,461   & 27,620    &   60,081   \\ 
	 \textsc{CrisisLexT26}  &   16,849    & 7,731    &   24,580  \\
	 \textsc{CrisisMMD}       &  11,509 &  4,549 & 16,058 \\  
	 %- - - Train &  8285  & 3299 &  11584 \\
	 % - - - Val   & 1612 & 625 & 2237  \\
     % - - - Test  & 1612 & 625 & 2237   \\
     % - - - TOTAL  & 11509 &  4549 & 16058\footnote{does not contain duplicates as the original dataset}\\
      \bottomrule   
\end{tabular}
%\vspace{-3mm}
\end{table*}

\begin{table*}[t]
%\vspace{-1.5mm}
\centering
\caption{Subsets classes distribution.}
%\vspace{-2.5mm}
\label{tbl:datasetsstatsSubset}
\small
\begin{tabular}{|l|r|r|r|}	
     \toprule
     \multicolumn{1}{|c|}{\textit{Dataset}}    & \multicolumn{1}{|c|}{\textit{\#Informative}}    & \multicolumn{1}{|c|}{\textit{\#Not Informative}}     & \multicolumn{1}{|c|}{\textit{Total}}   \\ 
     \midrule
     \textsc{Covid}$_{SUBSET}$      &    3,378 & 3,816 & 7,194    \\
	  %- - - Train &  &   &   \\
	  %- - - Val   &  &   &  \\
      %- - - Test  &  944 & 1056  & 2000\\ labelled test has not being released yet
      %- - - TOTAL  & 3378 & 3816 & 7194  \footnote{delete duplicates} \footnote{Missing the labelled test set}\\
     \textsc{CrisisLexT6}$_{SUBSET}$  &   20,568  & 17,422  & 37,990 \\ 
	 \textsc{CrisisLexT26}$_{SUBSET}$   &  11,023     &  4,442 &  15,465  \\
	 \textsc{CrisisMMD}$_{SUBSET}$          & 9,343 & 3,443 & 12,786  \\  
	 %- - - Train &   & &  \\
	 % - - - Val   &  & &  \\
     % - - - Test  &  & &   \\
      %- - - TOTAL  & 9343 & 3443 & 12786 \footnote{does not contain duplicates as the original dataset}\\
      \bottomrule   
\end{tabular}
%\vspace{-3mm}
\end{table*}

\subsection{Evaluation Metrics}
To evaluate the informative tweets classification task we employ the following performance metrics.
\textbf{Precision (P):} the fraction of the correctly classified instances among the instances assigned to the class.
\textbf{Recall (R):} the fraction of the correctly classified instances among all instances of the class and \textbf{F-score (F1):} the harmonic mean of precision and recall. In this work we compute the metrics independently for each class and then take the average, i.e. Macro Precision, Macro Recall and Macro F-score.

\subsection{Experiment settings}
\label{experiment_settings}
\vspace{-2mm}

We normalized all characters in the tweets to their lower-cased forms
followed by the removal of punctuation and non ASCII characters as well as non English words, then we calculated the text-based features and user-based features. The Bag-of-Words feature was calculated for the entire corpus of tweets, however in our experiments we only calculated it for words appearing at least 5 times in the entire corpus and up to a limit of 10000 times. The words with length less than two characters were also pruned. 

In parallel we then tokenized the sentences and encoded the tokens using the BERT encoder. Each dataset is randomly split into 10 mutually exclusive subets and 10-fold cross validation was used to measure the performance of the models. For the conventional classifiers we used the implementation from the scikit-learn tool \cite{pedregosa2011scikit} and all the  algorithms were set to use the default parameter values. As regards BERT fine tuning, we used the stochastic gradient descent optimizer with a learning rate of 0.001, momentum 0.9 and ran the training process for 20 epochs. We set the batch size to 16 and limited the BERT sentence encoding to the maximum length of 80. In this work the BERT models were built based on the pytorch-pretrained-BERT repository \url{https://github.com/huggingface/pytorch-pretrained-BERT.}

%\section{Results}
\subsection{\uppercase{Results}}
\label{results}
\vspace{-2mm}

We show the results in terms of macro average F-score.
Table~\ref{tbl:fulldetailed} shows the performance of the implemented models on all the datasets used in this work. The two best results obtained in each dataset is highlighted in bold face.
Only the \textsc{Covid} and \textsc{CrisisMMD} datasets were split into training and validation sets by default, however to make it fair and comparable across all the datasets and approaches we performed 10-fold cross validation with the entire datasets (combined training and validation sets).

In the first six rows we show the classification performance of conventional classifiers using the handcrafted features proposed in this work.
For the full datasets it is only possible to calculate the Twitter-based features, as the user-based features are strongly dependent on the complete tweet information, and since we had to crawl the Twitter platform to obtain the complete information, we realised that many tweets had been deleted.

We noticed the performance of the classifiers varies on a per dataset basis and classifiers  performed differently on each of the datasets.
For the \textsc{Covid} dataset we observed the \textsc{Logistic Regression} classifier performed the best with Macro F1 of 57.07, while for \textsc{CrisisLexT6} datasets 
and \textsc{CrisisLexT26} \textsc{MLP} showed the best score 75.56 and 68.10, respectively. And for \textsc{CrisisMMD}, \textsc{Random Forest} outperformed the other classifiers with a score of 55.85.

The following six rows show the classification performance using Bag-of-Words as input features.
%but now using as input the sentences encoded features using the BERT encoder.
Here again we noticed the performance of the classifiers varies on a per dataset basis, however we observed considerable performance improvement across all datasets which demonstrates that the bag-of-words is a stronger features encoding method than the handcrafted features approach only.

In the following six rows we show the results of the classification task using a combination of the handcrafted features with the Bag-of-Words features. It is interesting to observe that for the majority of the classifiers this combination does not produce improved results over  the \textsc{Covid} and the \textsc{CrisisLexT6} datasets. Only \textsc{Naive Bayes} demonstrated considerable improvement over the previous approach for the \textsc{Covid} dataset. However, all the classifiers demonstrated improvement in the \textsc{CrisisLexT26} dataset when compared to using the Bag-of-Words only approach, and for the \textsc{CrisisMMD} dataset again only \textsc{Naive Bayes} demonstrated improvement when compared to the previous approach.

The next six rows show the results of the conventional classifiers using  Fasttext  word embeddings. For the \textsc{Covid} and  \textsc{CrisisLexT6} datasets, \textsc{MLP} produced the best results, while for the \textsc{CrisisLexT26} and for the \textsc{CrisisMMD} datasets, \textsc{Logistic Regression} demonstrated the best macro F-score.
In the following six rows we can see the classification results using GloVe word embeddings. The performance results observed from the classifiers using this embedding technique seem to be similar to the Fasttext word embeddings varying not too much across datasets. 

In the following six rows we show the performance results of one approach in which we use the conventional classifiers using BERT encoded features combined with the handcrafted features.
We have not noticed improvements using this approach of combining BERT word embeddings with handcrafted features on the \textsc{Covid} and \textsc{CrisisMMD} datasets, however we observed some improvements in the \textsc{CrisisLexT6} and \textsc{CrisisLexT26} dataset for the majority of the classifiers.

Finally in the last row we show the results of our proposed approach \textsc{BERT}$_{Hyb}$. Our model outperforms all the previously cited methods across all datasets used in this work.
For \textsc{Covid} dataset it produced a macro F-score of 84.41 which is 2.5 percentage points improvement over the best result from previous approaches (\textsc{LR} using Bag-of-Words features). For \textsc{CrisisLexT6} we observed 95.96 macro F-score, for \textsc{CrisisLexT26} we obtained 79.09 macro F-score, which is the highest improvement (7 percentage points over \textsc{SVM} using handcrafted features combined with Bag-of-Words) and for \textsc{CrisisMMD} our model produced 77.66 macro F-score.

There are some reasons that can explain why our hybrid model performs much better than other models tested in this paper.
The first one is the fact that BERT encoder uses a contextual representation in which it processes words in relation to all the other words in the sequence, rather than one by one separately, and the second reason is the fact that we ran several training iterations while adjusting weights, and using different optimization functions to minimise the training loss. 

\begin{table*}[h]
\vspace{-2mm}
\caption{Models performance on the original datasets.}
\centering
\renewcommand{\arraystretch}{0.7}
\setlength{\tabcolsep}{3.5pt}
\label{tbl:fulldetailed}
\scalebox{0.6}{ 
\begin{tabular}{@{}llcccccccccccc@{}}
\toprule
\multirow{3}{*}{\textsc{\textbf{Features}}} & \multirow{3}{*}{\textsc{\textbf{Models}}}     & \multicolumn{3}{c}{\textsc{Covid}}
    & \multicolumn{3}{c}{\textsc{CrisisLexT6}} 
    & \multicolumn{3}{c}{\textsc{CrisisLexT26}} 
    & \multicolumn{3}{c}{\textsc{CrisisMMD}} \\ \cmidrule(l){3-5} \cmidrule(l){6-8} \cmidrule(l){9-11} \cmidrule(l){12-14}

\multicolumn{1}{c}{} & \multicolumn{1}{c}{}  &  & & MacroF1 &  &   & MacroF1 &   &   & MacroF1 &  &   & MacroF1   \\ \midrule
\multirow{5}{*}{\textsc{Handcrafted}} 		
& \textsc{LR}           &		&		&	 57.07(+/- 0.02)	&		&		&	 75.09(+/- 0.14)		&		&	    &		 64.60(+/- 0.05)	&		&		&	 48.91(+/- 0.02)	\\
& \textsc{DT}      	    &		&		&	 51.99(+/- 0.02)	&		&		&	 72.39(+/- 0.14)		&		&		&		 61.82(+/- 0.05)	&		&		&	 55.63(+/- 0.02)	\\
&	\textsc{RF}     	&		&		&	 54.14(+/- 0.02)	&		&		&	 74.05(+/- 0.14)		&		&		&		 64.14(+/- 0.05)	&		&		&	 55.85(+/- 0.02)	\\
& \textsc{NB}       	&	 	&		&	 42.79(+/- 0.02)	&		&		&	 72.51(+/- 0.14)		&		&		&		 65.79(+/- 0.05)	&		&		&	 50.60(+/- 0.02)	\\
& \textsc{MLP}       	&		&		&	 49.84(+/- 0.02)	&		&		&	 75.56(+/- 0.14)		&		&		&		 68.10(+/- 0.05)	&		&		&	 48.42(+/- 0.02)	\\
&  \textsc{SVM}     	&		&		&	 56.11(+/- 0.02)	&		&		&	 75.41(+/- 0.14)		&		&		&		 65.53(+/- 0.05)	&		&		&	 49.81(+/- 0.02)	\\

  \midrule
  \multirow{5}{*}{\textsc{Bag-of-Words}} 																	
 & \textsc{LR}        &   &   & \textbf{81.90}(+/- 0.04) &   &   & 92.90(+/- 0.09) &   &  & 66.46(+/- 0.17) &   &   & 72.68(+/- 0.03) \\  
 & \textsc{DT}        &   &   & 75.13(+/- 0.04) &   &   & 91.42(+/- 0.09) &   &  & 53.04(+/- 0.17) &   &   & 68.97(+/- 0.03) \\
 & \textsc{RF}        &   &   & 81.06(+/- 0.04) &   &   & \textbf{93.51}(+/- 0.09) &   &  & 62.59(+/- 0.17) &   &   & 73.21(+/- 0.03) \\
 & \textsc{NB}          &   &    & 66.75(+/- 0.04) &    &    & 80.35(+/- 0.09) &   &  & 57.09(+/- 0.17) &   &    & 47.56(+/- 0.03) \\
 & \textsc{MLP}         &   &    & 75.23(+/- 0.04) &    &    & 91.74(+/- 0.09) &   &  & 63.39(+/- 0.17) &   &    & 71.48(+/- 0.03) \\
 & \textsc{SVM}         &   &    & 81.38(+/- 0.04) &    &    & 93.21(+/- 0.09) &   &  & 65.01(+/- 0.17) &   &    & 66.00(+/- 0.03) \\
  \midrule

\multirow{5}{*}{\textsc{Handcrafted + BoW}} 		
 & \textsc{LR}   &  &   & 78.29(+/- 0.05) &  &  & 83.58(+/- 0.12) &   &  & 69.70(+/- 0.12) & &  & 65.12(+/- 0.03)\\ 
 & \textsc{DT}   &  &   & 74.68(+/- 0.05) &  &  & 90.80(+/- 0.12) &   &  & 61.26(+/- 0.12) & &  & 66.24(+/- 0.03)\\
 & \textsc{RF}   &  &   & 80.47(+/- 0.05) &  &  & 93.28(+/- 0.12) &   &  & 66.55(+/- 0.12) & &  & 70.61(+/- 0.03)\\
 & \textsc{NB}   &  &   & 71.56(+/- 0.05) &  &  & 79.00(+/- 0.12) &   &  & 60.83(+/- 0.12) & &  & 57.53(+/- 0.03) \\ 
 & \textsc{MLP}  &  &   & 75.28(+/- 0.05) &  &  & 91.51(+/- 0.12) &   &  & 63.96(+/- 0.12) & &  & 69.58(+/- 0.03)\\
 & \textsc{SVM}  &  &   & 75.05(+/- 0.05) &  &  & 92.96(+/- 0.12) &   &  & \textbf{72.09}(+/- 0.12) & &  & 66.18(+/- 0.03)\\
 \midrule
\multirow{5}{*}{\textsc{Fasttext}}
&  \textsc{LR}  	&	&	&	 77.60(+/- 0.04) & 	& 	&	89.29(+/- 0.08)	&	&	&  71.30(+/- 0.10)	&	&	& 74.09(+/- 0.02)	\\
&  \textsc{DT}  	&	&	&	 64.42(+/- 0.04) &	&	&	79.26(+/- 0.08)	&	&	&  60.85(+/- 0.10)	& 	&	& 63.74(+/- 0.02)	\\
&	\textsc{RF}    	&	&	&	 76.16(+/- 0.04) & 	&	&	88.62(+/- 0.08)	& 	&	&  69.83(+/- 0.10)	&	&	& 71.54(+/- 0.02)	\\
&	\textsc{NB}   	&   &   &	 74.73(+/- 0.04) &  &	&	77.18(+/- 0.08)	&	&	&  63.41(+/- 0.10)	&	&	& 66.89(+/- 0.02)	\\
&	\textsc{MLP}   	& 	&	&	 80.01(+/- 0.04) &  & 	&	91.28(+/- 0.08)	&	&	&  67.81(+/- 0.10)	&	&	& 74.00(+/- 0.02)	\\
&	\textsc{SVM}   	&	&	&	 76.43(+/- 0.04) &	&	&	89.46(+/- 0.08)	&	&	&  70.14(+/- 0.10)	&	& 	& 71.40(+/- 0.02)	\\
\midrule
  \multirow{5}{*}{\textsc{GloVe}}
&	\textsc{LR} 	  &   & 	&  79.68(+/- 0.04)	& &	 &	86.82(+/- 0.11)	&	& &	 70.40(+/- 0.09) & 	& &	 74.59(+/- 0.02)	\\
&	\textsc{DT}       &	  &	  	&  66.76(+/- 0.04)	& &	 &	77.49(+/- 0.11)	&	& &  60.05(+/- 0.09) &	& &	 63.54(+/- 0.02)	\\
&  \textsc{RF}        &   &  	&  77.80(+/- 0.04)	& &	 &  87.36(+/- 0.11)	&	& &	 66.27(+/- 0.09) & 	& &	 72.41(+/- 0.02)	\\
&\textsc{NB}          &	  &		&  76.29(+/- 0.04)	& &	 &  81.72(+/- 0.11)	&	& &	 61.30(+/- 0.09) & 	& &  72.87(+/- 0.02)	\\
&\textsc{MLP}         &	  &	  	&  79.03(+/- 0.04)	& &	 &	87.96(+/- 0.11)	&	& &	 66.21(+/- 0.09) &  & &	 72.38(+/- 0.02)	\\
&\textsc{SVM}         &   &	    &  80.05(+/- 0.04)	& &	 &  89.20(+/- 0.11)	&	& &	 71.75(+/- 0.09) & 	& &	 \textbf{75.15}(+/- 0.02)	\\
\midrule
 \multirow{5}{*}{\textsc{BERT} } 
 & \textsc{LR} &   &  & 77.83(+/- 0.03) &  &  & 90.62(+/- 0.09) &  &  & 70.41(+/- 0.10) &  &  & 74.80(+/- 0.03) \\
 & \textsc{DT} &   &  & 62.19(+/- 0.03) &  &  & 77.84(+/- 0.09) &  &  & 60.98(+/- 0.10) &  &  & 62.84(+/- 0.03) \\
 & \textsc{RF} &   &  & 74.11(+/- 0.03) &  &  & 87.51(+/- 0.09) &  &  & 69.11(+/- 0.10) &  &  & 70.76(+/- 0.03) \\
 & \textsc{NB} &   &  & 71.34(+/- 0.03) &  &  & 77.69(+/- 0.09) &  &  & 67.59(+/- 0.10) &  &  & 70.41(+/- 0.03) \\
 & \textsc{MLP}&   &  & 77.08(+/- 0.03) &  &  & 89.75(+/- 0.09) &  &  & 66.54(+/- 0.10) &  &  & 72.21(+/- 0.03) \\
 & \textsc{SVM}&   &  & 78.08(+/- 0.03) &  &  & 91.50(+/- 0.09) &  &  & 70.53(+/- 0.10) &  &  & 75.14(+/- 0.03) \\           
  \midrule
  %\textsc{BERT}$_{SEQ}$ &  & 83.16 & 82.97 &	83.01(+/- 0.01) & 93.94 & 93.22 & 93.48 (+/- 0.03) &  80.00 & 76.41 &	77.59 (+/- 0.04) & 80.81 & 75.19 & 77.09	(+/- 0.01)\\
  %\midrule
  %\textsc{BERT}$_{Hyb}$ & \textit{10epochs} & 81.13 & 81.05 & 80.80(+/- 0.01) &\\
  \midrule
  \textsc{Handcrafted + BERT} & \textsc{BERT}$_{Hyb}$ &   &	 &	\textbf{84.41}(+/- 0.01) &   &  	& \textbf{95.96}(+/- 0.03) &   &  & \textbf{79.09}(+/- 	0.04) &  &	& \textbf{77.66}(+/- 0.01) \\
  \bottomrule
\end{tabular}
}
\vspace{-2mm}
\end{table*}											

%\subsection{Subsets Evaluation}
We also evaluated the proposed approach in the subsets of the original datasets. As mentioned before these subsets were created so we could also calculate features related to the user who posted the message. We noticed again that the handcrafted features alone did not produce satisfactory results. The best observed macro F-scores   varied between 55.49 for the \textsc{CrisisMMD}$_{SUBSET}$ using a \textsc{Naive Bayes} classifier and 78.58 for the \textsc{CrisisLexT6}$_{SUBSET}$ using \textsc{Random Forest} classifier.
However when we used the Bag-of-Words model as input features, the classifiers  produced considerably better results for \textsc{Covid$_{SUBSET}$}  and \textsc{CrisisLexT6}$_{SUBSET}$ datasets in all cases, but for the \textsc{CrisisLexT26}$_{SUBSET}$ and \textsc{CrisisMMD}$_{SUBSET}$ there were some classifiers that performed better using only the handcrafted features, for example for the  \textsc{CrisisLexT26}$_{SUBSET}$ the \textsc{Random Forest} model produced a macro F-score of 66.80, while using the Bag-of-Words model it produced only 52.06. The combination of the handcrafted features and Bag-of-Words shows improvement for all datasets only when using the \textsc{Naive Bayes} classifier when compared to the Bag-of-Words model, while when compared to the sole handcrafted features the classifiers produce better results in all cases for the \textsc{Covid$_{SUBSET}$} and \textsc{CrisisLexT6}$_{SUBSET}$ datasets and the majority of cases in \textsc{CrisisLexT26}$_{SUBSET}$ and \textsc{CrisisMMD}$_{SUBSET}$ with the exception of the \textsc{Naive Bayes} classifier.

Using the Fasttext, GloVe and BERT embeddings as input features to the conventional classifiers showed considerable improvements across all datasets, especially when using \textsc{Logistic Regression} as base classifier, however this was not a pattern observed when using different classification methods.

Our hybrid model \textsc{BERT}$_{Hyb}$ produced the best performance result for almost all the dataset with the exception of the \textsc{CrisisLexT6}$_{SUBSET}$, however the difference is marginal. The best observed macro F-score is shown when using the Bag-of-Words features model using \textsc{Random Forest} as base classifier (93.22), while our hybrid approach produced a score of 93.05. In the \textsc{Covid$_{SUBSET}$} our model showed 84.64 macho F-score which is 2.3 percentage points improvement over the second best result (Bag-of-Words and \textsc{LR} = 82.35).
Our model showed 76.68 and 76.54  macro F-score for the \textsc{CrisisLexT26}$_{SUBSET}$ and  \textsc{CrisisMMD}$_{SUBSET}$ datasets respectively. These two datasets seem to be the two datasets where the performance of the models were lower than 80\%. Further investigation and a more in depth analysis is required as there is still some room for improvements.

\section{\uppercase{Conclusions}}
\label{Conclusion}
\vspace{-2mm}

Social media has drawn attention from different sectors of society and the information available during catastrophic events is extremely useful for both the ordinary citizen and the professionals involved in humanitarian purposes, however there is an overload of information that requires an automated filtering method for real time processing of relevant content. 

In this work we designed a set of handcrafted features from both the Twitter posts and the users who posted a tweet, and showed experimentally the performance of six conventional classifiers on the informative tweet classification task.  We also trained classifiers with several word embeddings, namely, Fasttext, GloVe and BERT, as input features.  Moreover, we showed that our proposed deep neural model \textsc{BERT}$_{Hyb}$ is more effective in identifying informative tweets as compared to conventional classifiers in different crisis related corpus from Twitter.

As future works we intend to further investigate different deep learning models combinations and implement a complete pipeline where the tweets are crawled and classified in real time based on crisis related trending topics.

%\vfill
%\section*{\uppercase{Acknowledgements}}

%If any, should be placed before the references section
%without numbering. To do so please use the following command:
%\textit{$\backslash$section*\{ACKNOWLEDGEMENTS\}}

\bibliographystyle{apalike}
{\small
\bibliography{main}}

\end{document}